\pdfoutput=1

\documentclass[11pt]{article}

\usepackage[final]{acl}

\usepackage{times}
\usepackage{latexsym}

\usepackage[T1]{fontenc}

\usepackage[utf8]{inputenc}

\usepackage{microtype}

\usepackage{inconsolata}

\usepackage{graphicx}
\usepackage{tikz}
\usetikzlibrary{shapes.geometric, positioning, fit}

\usepackage{pifont}
\usepackage{stackengine}
\usepackage{multirow}
\usepackage{comment}
\usepackage{cuted,tcolorbox,lipsum} 
\usepackage{xspace}
\usepackage{booktabs}
\newcommand{\dataname}{MEMERAG\xspace}
\newcommand{\datanameext}{MEMERAG-Ext\xspace}
\newcommand{\gptfomini}{GPT-4o mini\xspace}
\newcommand{\qwenthrirtytwob}{Qwen 2.5 32B\xspace}
\usepackage{float}
\usepackage{placeins}
\usepackage{afterpage}
\usepackage{enumitem}
\usepackage{booktabs,siunitx}
\usepackage{colortbl}  
\usepackage{xcolor}    
\usepackage{tabularx} 

\usepackage{hyperref}
\usepackage{soul}
\title{\dataname: A Multilingual End-to-End Meta-Evaluation Benchmark for Retrieval Augmented Generation}

\author{
    \textbf{Mar\'ia Andrea Cruz Bland\'on\textsuperscript{1}}\thanks{Work done by first author during an internship at Amazon. First two authors contributed equally to this work. Corresponding author: Jayasimha Talur email \href{mailto:talurj@amazon.com}{talurj@amazon.com}.},
    \textbf{Jayasimha Talur}$^{*}$
    \textbf{Bruno Charron}, 
\\
    \textbf{Dong Liu},
    \textbf{Saab Mansour},
    \textbf{Marcello Federico}
\\
    \textsuperscript{1}Tampere University,
    Amazon
}

\graphicspath{ {./plots/} }
\usepackage{xcolor}

\def\showcomments{}  
\ifdefined\showcomments
    \newcommand{\sm}[1]{\textcolor{cyan}{$_{Saab}${[#1]}}}
    \newcommand{\bc}[1]{\textcolor{magenta}{$_{Bruno}${[#1]}}}
    \newcommand{\dl}[1]{\textcolor{orange}{$_{Dong}${[#1]}}}
    \newcommand{\jt}[1]{\textcolor{red}{$_{Jayasimha}${[#1]}}}
    \newcommand{\mf}[1]{\textcolor{red}{$_{MF}${[#1]}}}
    
\else
    \newcommand{\sm}[1]{}
    \newcommand{\bc}[1]{}
    \newcommand{\dl}[1]{}
    \newcommand{\jt}[1]{}
    \newcommand{\mf}[1]{}
\fi

\newif\ifsectionenabled
\sectionenabledfalse

\newcommand{\best}[1]{{\textbf{#1}}}
\newcommand{\comparable}[1]{#1{$^{\dagger}$}}

\usepackage{setspace}

\begin{document}
\maketitle
\begin{abstract}
Automatic evaluation of retrieval augmented generation (RAG) systems relies on fine-grained dimensions like faithfulness and relevance, as judged by expert human annotators. Meta-evaluation benchmarks support the development of automatic evaluators that correlate well with human judgement. However, existing benchmarks predominantly focus on English or use translated data, which fails to capture cultural nuances. A native approach provides a better representation of the end user experience.

In this work, we develop a Multilingual End-to-end Meta-Evaluation RAG benchmark (\dataname).
Our benchmark builds on the popular MIRACL dataset, using native-language questions and generating responses with diverse large language models (LLMs), which are then assessed by expert annotators for faithfulness and relevance. We describe our annotation process and show that it achieves high inter-annotator agreement. We then analyse the performance of the answer-generating LLMs across languages as per the human evaluators. Finally we apply the dataset to our main use-case which is to benchmark multilingual automatic evaluators (LLM-as-a-judge). We show that our benchmark can reliably identify improvements offered by advanced prompting techniques and LLMs~\footnote{We release~\url{https://github.com/amazon-science/MEMERAG} our benchmark to support the community developing accurate evaluation methods for multilingual RAG systems.}. 
\end{abstract}

\section{Introduction}
Retrieval augmented generation (RAG) is emerging as a popular application of large language models (LLMs) and a powerful paradigm to improve LLMs factuality \cite{gao_retrieval-augmented_2024}. A RAG pipeline first retrieves relevant documents from an index based on a query and then composes a response using an LLM. Grounding the LLM response on retrieved knowledge helps mitigate outdated knowledge, lack of domain expertise and reduce hallucinations \cite{lewis2020retrieval, gao_retrieval-augmented_2024}.
Collecting benchmarking data for RAG is challenging due to the complexity of the pipeline that includes {\em information retrieval} and {\em text generation}. Text generation, our focus in this paper, has, in general,  two modes of {\em automatic evaluation}: reference-based and reference-free, which differ in the availability of human-generated gold references for each model input. Both modes can either leverage single (e.g. BERTScore \cite{Zhang2020BERTScore:}) or multidimensional (e.g. autoMQM  \cite{fernandes_devil_2023}) scores. Multidimensional evaluation 
\cite{burchardt-2013-mqm} provides more comprehensive understanding of text generation systems and is the de facto 
standard in the machine translation (MT) community\footnote{Since 2021 in the WMT metrics shared task \url{https://www2.statmt.org/wmt24/metrics-task.html}}. 

In a reference-free evaluation setup,  gold multidimensional judgements (factuality, relevance, etc) of model generations can be leveraged in a {\em meta-evaluation} framework. In this framework, automated evaluators are evaluated against human judgement to measure correlation. The automated evaluators can then be applied to measure the performance of new models' output. 

Previous work for RAG meta-evaluation mainly focused on English \cite{fan-2024-rag-survey} or leveraged human or machine translation of English datasets \cite{sharma2024fauxpolyglotstudyinformation}. Multilingual meta-evaluation is important to reliably measure performance across languages which can vary depending on language characteristics (low vs. high resource, complex morphology, etc.) and scripts (Latin vs. non-Latin). Translation-based benchmarks, while permitting cross-language comparisons,  suffer from translationese phenomena such as introducing simpler syntax and  lexical choices \cite{Baker1993CorpusLA, graham-etal-2020-statistical}, thus leading to data distributionally different from native data and not necessarily reflecting native users preferences \cite{chen-etal-2024-good-data}. Our position is that translation-based (parallel) benchmarks should be complemented by native multilingual benchmarks. 

To bridge those gaps, we propose a native meta-evaluation multilingual benchmark for RAG systems. Our benchmark is built on top of the popular MIRACL \cite{zhang_miracl_2023} dataset\footnote{\url{https://huggingface.co/datasets/miracl/miracl}} that includes native questions across 18 languages and relevance judgements of retrieved passages for multilingual retrieval evaluation. We extend MIRACL by generating answers in five languages with a diverse set of LLMs, and collecting judgements on the faithfulness and relevance of the answers using native expert human annotators. For the latter, we devised a structured annotation process that achieved a high rate of inter-annotator agreement. To evaluate the benchmark and set reference baseline results for others to compare against, we run LLM-as-a-judge experiments with  various prompting techniques and state-of-the-art LLMs. 


To summarize, our main contributions are:
\begin{itemize}[noitemsep,topsep=0pt]
    \item We built and publicly release the first (to the best of our knowledge) native multilingual  meta-evaluation RAG benchmark.
    \item We developed a rigorous flow chart-based annotation process to achieve high inter-annotator agreement rate for both faithfulness and relevance judgements. 
    \item We evaluated the quality of the benchmark on three multi-lingual meta-evaluation aspects: prompt selection, model selection, and fine-grained analysis.
    \item We establish reference baselines of multilingual automatic evaluators on our benchmark, showcasing performance improvements when using advanced prompting and LLMs.
\end{itemize}

\begin{figure*}[!t]
  \centering
  \includegraphics[width=1\textwidth]{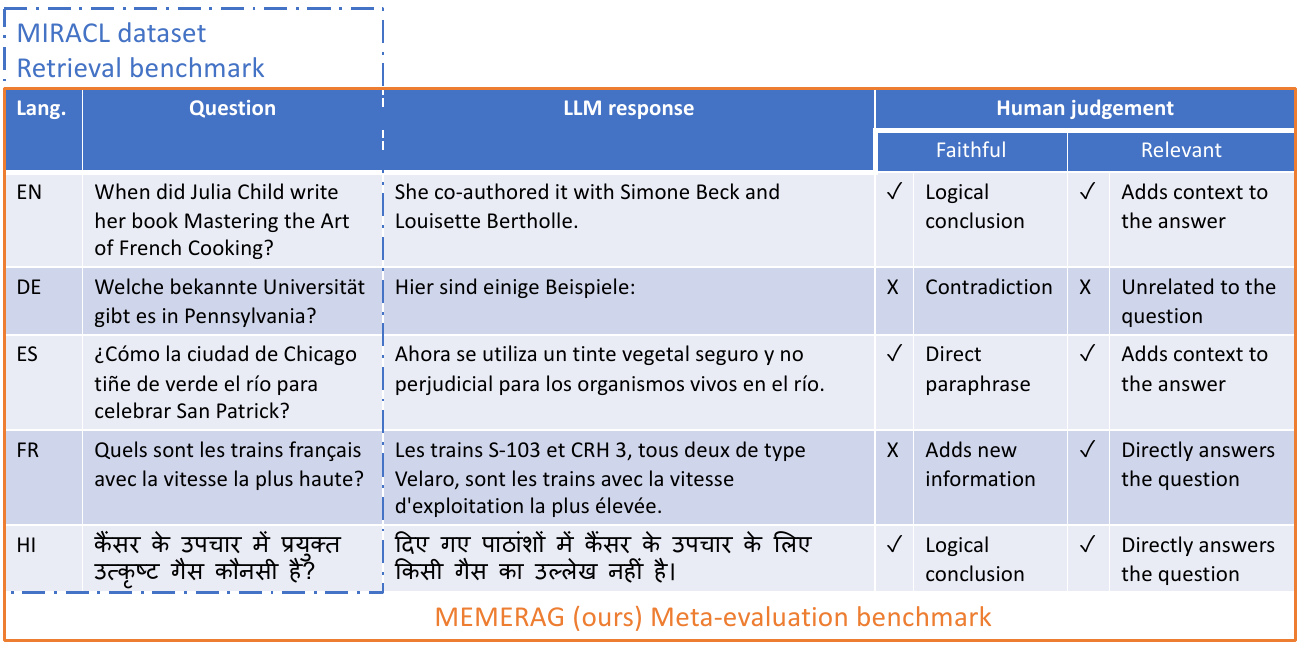} \caption{Examples from our Multilingual End-to-end Meta-Evaluation for RAG (MEMERAG) dataset. We select MIRACL native multilingual questions (for a subset of 5 languages), generate responses using diverse LLMs, and annotate each generated sentence with human experts for faithfulness and relevance. The annotation includes both coarse-grained (\ding{51}, \ding{53}) and fine-grained labels. Our dataset forms a meta-evaluation benchmark where automated evaluators can be developed and assessed on their correlation to human judgement. Due to space constraints we omit the retrieved documents (context) for the question and show only one sentence per LLM response. 
  }
  \label{fig:derive_dataset}
\end{figure*}
\section{Related Work}
Due to the pipeline approach of RAG systems, the evaluation can be split into 3 main components: 1) retriever metrics to identify relevant chunks of information to the input typically measured with recall/precision@K \cite{manning2008ir}; 2) generator metrics to identify the “usefulness” of the generated answers in relation to the input, this is done across fine grained dimensions such as faithfulness and relevance, either leveraging references answers \cite{es-etal-2024-ragas}; 3) end-to-end (overall) metrics that take into account additional components such as preprocessing, chunking, query reformulation and cascading errors. Retrieval metrics have been extensively studied in the information retrieval community, hence recent work focused on the text generation performance of RAG systems. This is also the focus of our work. One important difference between generation with and without retrieved documents is the conflict between the parametric “world knowledge” and the non-parametric retrieved documents knowledge, hence the distinction between faithfulness against the retrieved documents (RAG specific) and factuality according to general knowledge~\cite{maynez-etal-2020-faithfulness, wu2024clasheval}.


Recent studies have investigated the importance of various components within multilingual RAG systems. \cite{chirkova2024s} utilized existing multilingual QA datasets to evaluate different combinations of retrievers and generator models, finding that task-specific prompt engineering is crucial for high-quality multilingual generation. In another study, \cite{mirage_bench} extend MIRACL dataset to develop ``MIRAGE-BENCH'' a synthetic arena-based benchmark for ranking multilingual LLM generators. They employed preference judgments from a GPT-4o ``judge'' to train a ranking model. However, an important limitation of such synthetic benchmarks is the potential for self-preference bias \cite{self_preference_bias}, where the LLM judge may favor its own generations.

In this work, we focus on the faithfulness and relevance aspects to ensure meaningful results. These dimensions are typically assessed by human evaluators based on model-generated outputs, highlighting the need for developing automatic evaluation metrics that correlate well with human judgment—a process known as meta-evaluation. This need has led to a recent trend in the English-language research community of publishing meta-evaluation datasets and developing automated evaluators \cite{es-etal-2024-ragas, saad-falcon_ares_2024}.




To the best of our knowledge, no multilingual meta-evaluation benchmark for RAG systems currently exists. In this work, we address this gap by developing such a benchmark. Our dataset facilitates the creation of multilingual automatic evaluators that correlate well with human judgments. This, in turn, enables comprehensive end-to-end benchmarking of RAG systems. We believe our dataset is the first to offer this capability in a multilingual context.

\section{Dataset Construction}
\label{sec:dataset_construction}
Meta-evaluation datasets enable the development of reliable automatic evaluators. In a RAG setup, the input to the evaluator is composed of a question $q$ (user input), a context $c$ (set of passages automatically retrieved to the question) and an answer $a$ generated by a language model to answer the question based on the context. The end-to-end evaluator then needs to judge the quality of the answer $a$ given the context and the question $(c,q)$. 
Following previous work \cite{saad-falcon_ares_2024, es-etal-2024-ragas} we focus on two quality dimensions: 
\begin{center}
\noindent\fbox{\parbox[t]{0.96\linewidth}{%
\setlength{\parskip}{0pt}%
\noindent\textbf{Faithfulness } Is the answer grounded on the context, regardless of your world knowledge?\\[0.5em]
\noindent\textbf{Relevance } Is the answer relevant to the question, regardless of the context?
}}
\end{center}

We build a multilingual end-to-end meta-evaluation RAG (\dataname) dataset by extending the MIRACL dataset~\citep{zhang_miracl_2023} to include \textit{model-generated answers and human-based quality judgements}. 
More precisely, we select relevant question-context pairs, generate answers using various language models and gather expert human annotations on the quality of those answers.
Our dataset encompasses 5 languages: English (EN), German (DE), Spanish (ES), French (FR), and Hindi (HI), which represent multiple language families and both high- and low-resource languages.
Figure~\ref{fig:derive_dataset} shows examples from the dataset, with LLM-generated answers and coarse- to fine-grained human-assigned labels for the faithfulness and relevance dimensions. 

The MIRACL dataset is composed of questions written by humans in their \textit{native} languages, one or more passages automatically retrieved from the Wikipedia, and human annotations about the relevance of each passage.
Building a dataset starting from native questions in each language allows to evaluate RAG pipelines without resorting to (machine) translations, thus avoiding limitations and biases associated with translation. 
Note, however, that as questions were elicited from native speakers independently across different languages, the resulting data set is not parallel.

\subsection{Question Selection}
The questions in the MIRACL dataset were generated by humans based on prompts. 
This leads to questions that may be answerable by the prompt but may have ambiguity outside of that context.
In particular, we identified as problematic the questions for which the right answer can change over time.
For example ``\textit{Who is the president of Spain?}''
or ``\textit{How old is Drake Hogestyn?}''.
In a RAG setting, different passages may have been written at different times and provide conflicting context.
Additionally, the time of reference is usually not explicit in the question.
To remove those complications we automatically filtered out time-dependent questions across all languages\footnote{
See Appendix~\ref{sec:prompts_dataset_construction} for all prompts used during dataset construction. 
}.
We combined the train and dev splits of the MIRACL dataset, which corresponds to a total of 3,662, 305, 2,810, 1,486 and 1,519 questions respectively for EN, DE, ES, FR and HI.
Of those, 244, 13, 120, 64 and 86 were identified as time-dependent and filtered out.

\subsection{Context Selection}

The MIRACL dataset has an average of 10.3 passages per question, which corresponds to an average of 1,218 words of context in the English train and dev splits, with similar numbers in other languages.
To reduce the cognitive load on human annotators, we limit the context per query to 5 passages.

The source dataset provides human-annotated binary relevance labels for passages. To more accurately simulate an automated retrieval process, we rank the passages for each question using BM25~\citep{schutze2008introduction}, as implemented in \cite{bm25s}. We then select the top-5 ranked passages for each question.
If these top-5 passages do not contain any human-annotated relevant passages, we replace the lowest-ranked passage with the highest-ranked relevant passage from the full set. This approach ensures that each question has at least one relevant passage in its context, avoiding scenarios where annotators would evaluate responses without any relevant information.

It is worth noting that simulating scenarios where no relevant passages exist is straightforward (e.g., by including only irrelevant passages). In such cases, for faithfulness evaluation, we would expect responses like "The provided documents do not contain a relevant answer." Our method focuses on faithfulness while efficiently utilizing human annotation efforts by ensuring that each evaluated case has at least some relevant context.

\subsection{Answer Generation}
\label{sec:answer_gen}

After question and passage selection, we generate an answer for each question-context pair and each of five state-of-the-art LLMs. We generated answers with Claude 3 Sonnet, Llama3 70B, Llama3 8B, Mistral 7B, and \gptfomini.
Those LLMs were selected to cover a range of model sizes, open weight and proprietary models.
We prompted all the models in English\footnote{There is evidence in the literature for better model accuracy when the models are prompted to process in English~\cite{lai-etal-2023-chatgpt, liu2024translationneedstudysolving}, though under particular scenarios other languages could perform better~\cite{behzad-etal-2024-ask}. We leave additional prompting experiments for future work.}, asking to answer the question based only on the given context, and requesting the answer to be provided in the same language as the context and question.
For all models, we set the temperature to $0.1$, and maximum number of output tokens to 1000.

We thus produced answers for more than 1000 questions per language, except for German for which MIRACL only contains 305 questions. As our focus is on long-form answers, we further filtered out questions for which any of the 5 models generated an answer shorter than 10 words.

\subsection{Annotation Guidelines}

The task of annotating answers with faithfulness is challenging  due to several factors.
First, it involves some subjectivity which might impact Inter-Annotator Agreement (IAA) \citep{kryscinski_evaluating_2020, tang_tofueval_2024}.
Then,  its label space is not precisely defined in the literature \citep{tang_tofueval_2024,laban_summedits_2023, malaviya_expertqa_2024}. 
Finally, although faithfulness should be ideally evaluated for atomic facts, it is generally evaluated at the sentence or even document level, due to annotation costs.

Starting with the factuality error taxonomy introduced in~\citet{tang_tofueval_2024}, we ran a number of annotation pilots to refine the label space and guidelines.

Finally, we converged to three coarse-grained labels (\textit{Supported}, \textit{Not supported}, \textit{Challenging to determine}), explained through 10 fine-grained labels.
To increase the consistency of the annotation (IAA), we guide the annotation process through a flow chart (documented in Figure~\ref{fig:annotation_guide} in Appendix~\ref{appendix:annotation_guideline}).
For relevance, which is significantly less ambiguous to evaluate, we device a simple annotation process with three labels: \textit{Directly answers the question}, \textit{Adds context to the answer}, and \textit{Unrelated to the question}. Note that the first two labels can be used to describe "relevant" sentences, while the last label identifies "irrelevant" sentences. 
(See Appendix~\ref{appendix:annotation_guideline} for more details.)

\begin{table}
    \centering
    \normalsize{
    \begin{tabular}{lcccc}
    \hline
    \multirow{2}{*}{\textbf{Lang}} & \multirow{2}{*}{\textbf{\#Q}} & \multicolumn{2}{c}{\textbf{Answer}} & \textbf{Context} \\
    &  & \textbf{\#S} & \textbf{Avg. \#W}  & \textbf{Avg. \#W} \\ 
    \hline
    EN & 250 & 400 & 30.3 & 613.5 \\
    DE & 250 & 468 & 27.3 & 455.0 \\
    ES & 250 & 563 & 52.1 & 522.3 \\
    FR & 250 & 540 & 48.7 & 478.3 \\
    HI & 250 & 351 & 23.8 & 571.5 \\\hline
    Total & 1,250 & 2,322 & & \\\hline
  \end{tabular}
  }
    \caption{General statistics of the \dataname dataset. \#Q: number of questions, \#S: number of sentences, \#W: number of words.
    Each answer is annotated at the sentence level, leading to 2,322 total sentences annotated by experts for faithfulness and relevance.
    }
    \label{tab:dataset_stats}
\end{table}

\subsection{Annotation Process}
\label{sec:annotations_process}

From the question-context-answer triplets obtained in Section~\ref{sec:answer_gen}, we randomly sampled 250 questions per language (50 per answer-generating model, without overlapping questions for diversity).
We employed a professional vendor with native annotators\footnote{Annotators were  compensated with a competitive hourly rate that is benchmarked against similar
roles in their country of residence.} to gather annotations for each sentence \footnote{Sentences were segmented using the pySBD\cite{sadvilkar-neumann-2020-pysbd} package.} of the 250 answers per language. The statistics of the annotated dataset are presented in Table~\ref{tab:dataset_stats}.
Among the 250 answers per language, a random subset of 10 were assigned to 3 annotators for computing the IAA and the rest to a single annotator (see Table~\ref{tab:iaa}). 

The annotations were gathered via a web-based tool that implemented the flow chart of the annotation guidelines (see Appendix~\ref{appendix:annotation_guideline} for details).
To further enhance IAA, we drew upon the findings of \citet{krishna-etal-2023-longeval}, which demonstrated that highlighting relevant information aids annotators in performing tasks and reaching consensus. Thus, we utilized the Llama 3 70B LLM to identify sentences within the retrieved passages that could potentially serve as supporting information to the answer sentences (see Appendix~\ref{sec:prompts_dataset_construction}).

The English annotations required approximately 25 hours of total annotation time, averaging 5.5 minutes per question. This covered 250 questions, including 10 that were annotated by three different annotators for quality control. Similar time investments were observed for the other four languages.

Table~\ref{tab:iaa} summarizes the IAA per language for faithfulness and relevance labels assigned by 3 annotators. 
We report IAA using Gwet's AC1~\cite{gwet2008computing} 
and Fleiss Kappa \cite{fleiss_kappa}.
We observe high agreement for faithfulness (0.84-0.93 Gwet's AC1 and 0.70-0.88 Fleiss Kappa) and even higher agreement for relevance (0.95-1.0 Gwet's AC1 and 0.63-1.0 Fleiss Kappa).
This shows that the annotators are aligned and indicates a high quality of the annotations. Comparing to previous work, \citet{tang_tofueval_2024} report a Fleiss Kappa of 0.34-0.42 on  faithfulness labels which they deem fair to moderate agreement. Note that a direct comparison with this work is not possible as they deal with different tasks, nevertheless the high IAA we are reporting is a testament of the effectiveness of our flow chart-based annotation design.
Further details on IAA with fine-grained explanatory labels, and an extended version of the dataset (\dataname-Ext) with five annotators for 150 answers per language can be found in Appendix~\ref{appendix:iaa_and_ext_dataset}.


\begin{table}[h]
\centering
\begin{tabular}{@{}ccccl@{}}
\toprule
\textbf{Lang} & \multicolumn{2}{c}{\textbf{Faithfulness}}              & \multicolumn{2}{c}{\textbf{Relevance}} \\ \midrule
              & \textbf{Gwet's} & \multicolumn{1}{l|}{\textbf{Fleiss}} & \textbf{Gwet's}    & \textbf{Fleiss}   \\
\multicolumn{1}{l}{} & \multicolumn{1}{l}{\textbf{AC1}} & \multicolumn{1}{l|}{\textbf{Kappa}} & \multicolumn{1}{l}{\textbf{AC1}} & \textbf{Kappa} \\ \midrule
EN            & 0.93            & \multicolumn{1}{c|}{0.77}            & 1.00               & 1.00               \\
DE            & 0.84            & \multicolumn{1}{c|}{0.81}            & 0.95               & 0.73              \\
ES            & 0.91            & \multicolumn{1}{c|}{0.76}            & 1.00               & 1.00               \\
FR            & 0.89            & \multicolumn{1}{c|}{0.88}            & 0.93               & 0.63              \\
HI            & 0.89            & \multicolumn{1}{c|}{0.70}             & 1.00               & 1.00               \\ \bottomrule
\end{tabular}
\caption{Inter-annotator agreement (IAA) on the  faithfulness and  relevance dimensions with 3 annotators. Annotations are at the answer sentence level.}
\label{tab:iaa}
\end{table}

\section{Annotation Results}
\label{sec:details-on-dataset}
We present in this section the results of the human annotations for the 2,322 sentences of the \dataname dataset.


\begin{table}[htbp]
    \centering
\begin{tabular}{lccc|ccc}
\toprule
\textbf{Lang} & \multicolumn{3}{c|}{\textbf{Faithfulness}} & \multicolumn{3}{c}{\textbf{Relevance}} \\
    & \ding{52} & \ding{53} & \multicolumn{1}{c|}{$\text{\sffamily ?}$} & \ding{52} & \ding{51} & \ding{53} \\
\midrule
EN & 65.2 & 31.5 & 3.2 & 65.2 & 32.5 & 2.2 \\
DE & 71.2 & 26.7 & 2.1 & 61.3 & 26.5 & 12.2 \\
ES & 65.7 & 32.9 & 1.4 & 48.8 & 43.9 & 7.3 \\
FR & 62.0 & 37.8 & 0.2 & 63.3 & 29.3 & 7.4 \\
HI & 73.8 & 25.6 & 0.6 & 68.9 & 21.4 & 9.7 \\
\bottomrule
\end{tabular}
    \caption{Label distribution in the benchmark. Percentage of sentences labelled as supported (\ding{52}), not supported (\ding{53}), challenging to determine ($\text{\sffamily ?}$) for  faithfulness, and as directly answers the question (\ding{52}), adds context to the answer (\ding{51}), unrelated to the question (\ding{53}) for relevance, per language.
    \label{tab:factuality}
    }
\end{table}
\noindent
Table~\ref{tab:factuality} shows the distribution of faithfulness and relevance labels across the five languages.
The distribution of labels is consistent across all languages, with a few exceptions.
On faithfulness, German and Hindi show higher percentages of \textit{Supported} answers.
On relevance, Spanish presents a significantly higher percentage of labels \textit{Adds context to the answer} compared to other languages while English had a very low share of labels \textit{Unrelated to the question}.
As a partial explanation for this, we note that Spanish questions generated the largest numbers of output sentences with 52.1 words vs., for example, 30.3 words for English (see Table~\ref{tab:dataset_stats}).
Hence, we expect that the relevance statistics reflect the tendency of Spanish answers to be more verbose.

\begin{table}[t]
\centering
    \resizebox{\columnwidth}{!}{
\begin{tabular}{l*{5}{S[table-format=2.1]}}
\toprule
\textbf{Label} & {\textbf{EN}} & {\textbf{DE}} & {\textbf{ES}} & {\textbf{FR}} & {\textbf{HI}} \\
\midrule
Direct paraphrase & 8.5 & 41.7 & 25.0 & 33.7 & 40.7 \\
Logical conclusion & 41.2 & 28.2 & 30.4 & 28.0 & 5.7 \\
Other & 15.5 & 1.3 & 10.3 & 0.4 & 27.4 \\
\midrule
Adds new info & 7.0 & 9.6 & 16.0 & 15.0 & 14.8 \\
Contradiction & 4.5 & 11.3 & 8.3 & 5.9 & 7.1 \\
Mis-referencing & 1.5 & 3.0 & 2.3 & 3.7 & 0.3 \\
Nuance shift & 6.8 & 0.6 & 4.3 & 5.6 & 1.7 \\
Opinion as fact & 0.5 & 0.6 & 0.2 & 2.2 & 0.3 \\
Wrong reasoning & 10.0 & 0.6 & 1.4 & 1.9 & 0.3 \\
Other  & 1.2 & 0.9 & 0.4 & 3.5 & 1.1 \\
\midrule
Challeng. to determ. & 3.2 & 2.1 & 1.4 & 0.2 & 0.6 \\
\bottomrule
\end{tabular}
}
    \caption{Fine-grained faithfulness label distribution in the benchmark. Percentage of sentences with each label per language. The three sections correspond to the coarse-grained labels \textit{Supported/Not supported/Challenging to determine}.
    }
\label{tab:mistake_analysis}
\end{table}

\noindent
For a more granular insights into the annotations, we show in  Table~\ref{tab:mistake_analysis} the distribution of the explanatory fine-grained faithfulness labels for each language. As label names are quite self-explanatory, we refer for their precise meaning to Figure~\ref{fig:annotation_guide} in Appendix~\ref{appendix:annotation_guideline}.   Table~\ref{tab:mistake_analysis} is split into three blocks, respectively, addressing fine-grained labels for \textit{Supported} (top), \textit{Not supported} (middle), and \textit{Challenging to determine} (bottom) answers.  We observe remarkable differences across languages. 
For example, supported answers for English are prominently under form of a \textit{logical conclusion} from the context (41.2\%),  for German and Hindi from \textit{direct paraphrasing} of information in the context (41.7\% and 40.7\%), 
while for French and Spanish from a more balanced combination of the two reasons. 
On the side of unsupported answers,  the main mistake type in English is \textit{Wrong reasoning} (10\%), i.e. answers are non logical conclusions from the context, while this type or error is significantly rarer (under 2\%) for all other languages. The rate of  \textit{Adds new information} errors, a.k.a. hallucinations, ranges from 7\% for English to 16\% for Spanish. 
The observed cross-linguistic variations in the distribution of labels can be attributed to  the different nature of the questions and accuracy of the models across the 5 languages.

\section{Dataset Applications}
\label{sec:experimental}

The \dataname dataset is designed to support the development of reliable automatic evaluation methods.
For that purpose, we describe in this section how our dataset can be used as a benchmark to enable various meta-evaluation use cases.
We focus on two applications: 1) \textbf{Prompt selection}: The ability of our benchmark to effectively select prompts for automatic evaluation, 2) \textbf{Model selection}: The effectiveness of our benchmark to compare and select models. 
By concentrating on these aspects, we can evaluate the benchmark's utility as a comprehensive tool for multilingual model assessment.
While we provide reference baselines for each application, our focus is on showcasing the effectiveness of the benchmark rather than the underlying capabilities of the LLMs.

\subsection{Experimental Setup}

\paragraph{Benchmark Tasks}
Our benchmark is composed of multiple tasks defined by the annotation dimension and subset considered.
On the annotation dimension, we focus our experiments on the coarse-grained faithfulness dimension.
This dimension is more challenging than relevance as highlighted by the lower IAA, while retaining a high-enough IAA to make for a trustworthy benchmark.
We invite benchmark users to also experiment on the other dimensions provided by the dataset depending on their use case.
Note that for this experiment, we remove the sentences labelled as \textit{Challenging to determine} by human annotators.
On the annotation subset, we first distinguish the \textit{multilingual} task which uses the full dataset and the \textit{monolingual} task, which only considers a single language.
Those tasks can then be further broken down at the \textit{fine-grained} level by considering the subset of sentences with a certain fine-grained label.
Performance is evaluated with Balanced Accuracy (BAcc), with equal weights on each coarse-grained label and language.
We conduct significance testing using permutation tests~\cite{good2013permutation}, with further details in Appendix~\ref{sec:stat_test}.

\label{sec:prompting_strategies}

\paragraph{Reference Prompts}
To demonstrate how the benchmark can be used to select the appropriate prompt, we experiment with multiple prompting strategies from simple to advanced techniques, starting with zero-shot prompting, where LLMs directly classify statements as \textit{Supported} or \textit{Not supported}. We then implement chain-of-thought (COT) prompting \cite{chain_of_thought_jason_wei}, which incorporates an intermediate reasoning step. While these basic prompting strategies provide a good starting point for evaluation, our initial experiments revealed limitations in their ability to capture the nuanced requirements of our specific task. Without explicit guidelines in the prompt, automatic evaluators rely on their ``world knowledge'', which may not align with the specific requirements of the evaluation task. To overcome this, we add  instructions from the annotation guidelines (AG) in the prompt. 
Adding annotation guidelines provides clear criteria for what constitutes \textit{Supported} versus \textit{Not Supported} sentences, reducing ambiguity in the evaluation process. The various prompts are presented in Appendix~\ref{sec:eval_prompts}.

\paragraph{Reference Models}
We experiment with four LLMs with varying model sizes and capabilities: \gptfomini, \qwenthrirtytwob, and two versions of Llama 3.2 (11B and 90B)\footnote{In comparison to the LLMs selected for answer generation, see Section~\ref{sec:answer_gen}, we upgraded Llama from 3 to 3.2 as the context length of 8K tokens was not sufficient for all prompts. We picked GPT-4o mini as representative of proprietary models. Additionally, Mistral was excluded as it performed poorly in initial experiments.}. In case an LLM does not produce one of the required labels, we repeatedly prompt the LLM up to five times with temperature and \verb|top_p| equal to 0.1 to get a valid label. If the LLM fails to generate a label after five retries, we treat the datapoint as an error (wrong label).

\begin{table}[]
\centering
\begin{tabular}{
    l
    S[table-format = 2.1]
    S[table-format = 2.1]
    S[table-format = 2.1]
    S[table-format = 2.1]
}
\toprule
\textbf{Prompt} & \textbf{\small \shortstack{GPT-4o\\mini}}    & \textbf{\small \shortstack{Qwen\\2.5 32B}}    & \textbf{\small \shortstack{Llama\\3.2 90B}}    & \textbf{\small \shortstack{Llama\\ 3.2 11B}}    \\ \midrule
ZS & 59.7 & 66.7 & 58.0 & 55.4 \\
COT & 61.4 & 68.8 & 59.9 & \best{62.5} \\
AG & \comparable{71.6} & \best{72.6} & \comparable{62.8} & 57.9 \\
AG+COT & \best{71.7} & \comparable{71.8} & \best{64.4} & \comparable{61.6} \\
\bottomrule
\end{tabular}
\caption{Reference baselines for the multilingual task on coarse-grained faithfulness. Balanced accuracy (BAcc) of the automatic evaluators averaged across the 5 languages (EN, DE, ES, FR, HI) using zero-shot (ZS), chain-of-thought (COT), annotation guidelines (AG) and AG+COT prompting strategies. Bold indicates best performance for the column, $\dagger$ indicates results not statistically different from the best (p > 0.05). Additional results on monolingual tasks and standard errors can be found in Appendix~\ref{sec:stat_test}, Tables \ref{tab:avg_llm_bacc_with_se}-\ref{tab:metaeval_all}.}
\label{tab:avg_prompt_bacc}
\end{table}

\begin{figure*}[t]
  \centering\small
  \begin{tabular}{cc}
    \includegraphics[width=0.48\linewidth]{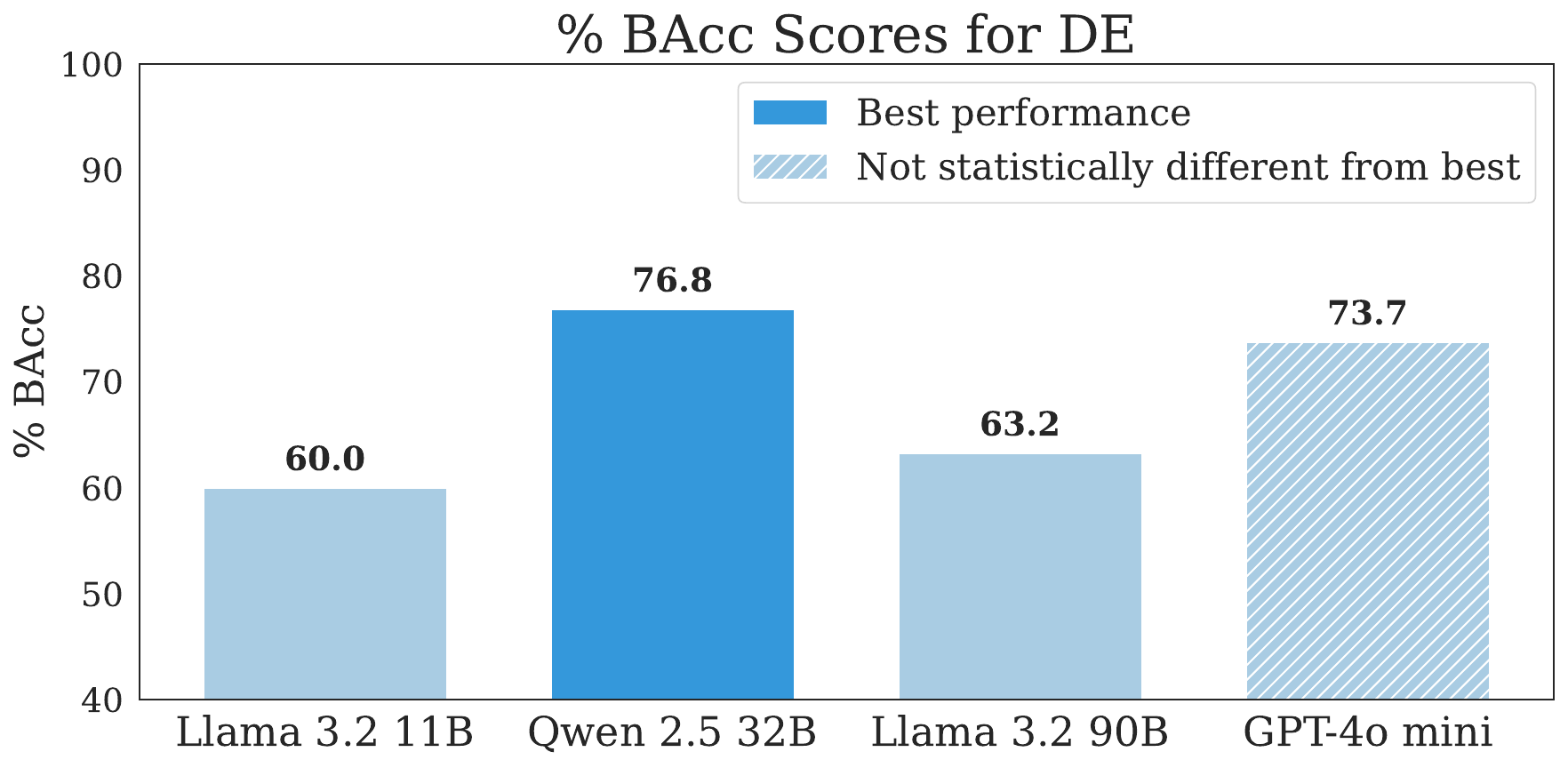} &
    \includegraphics[width=0.48\linewidth]{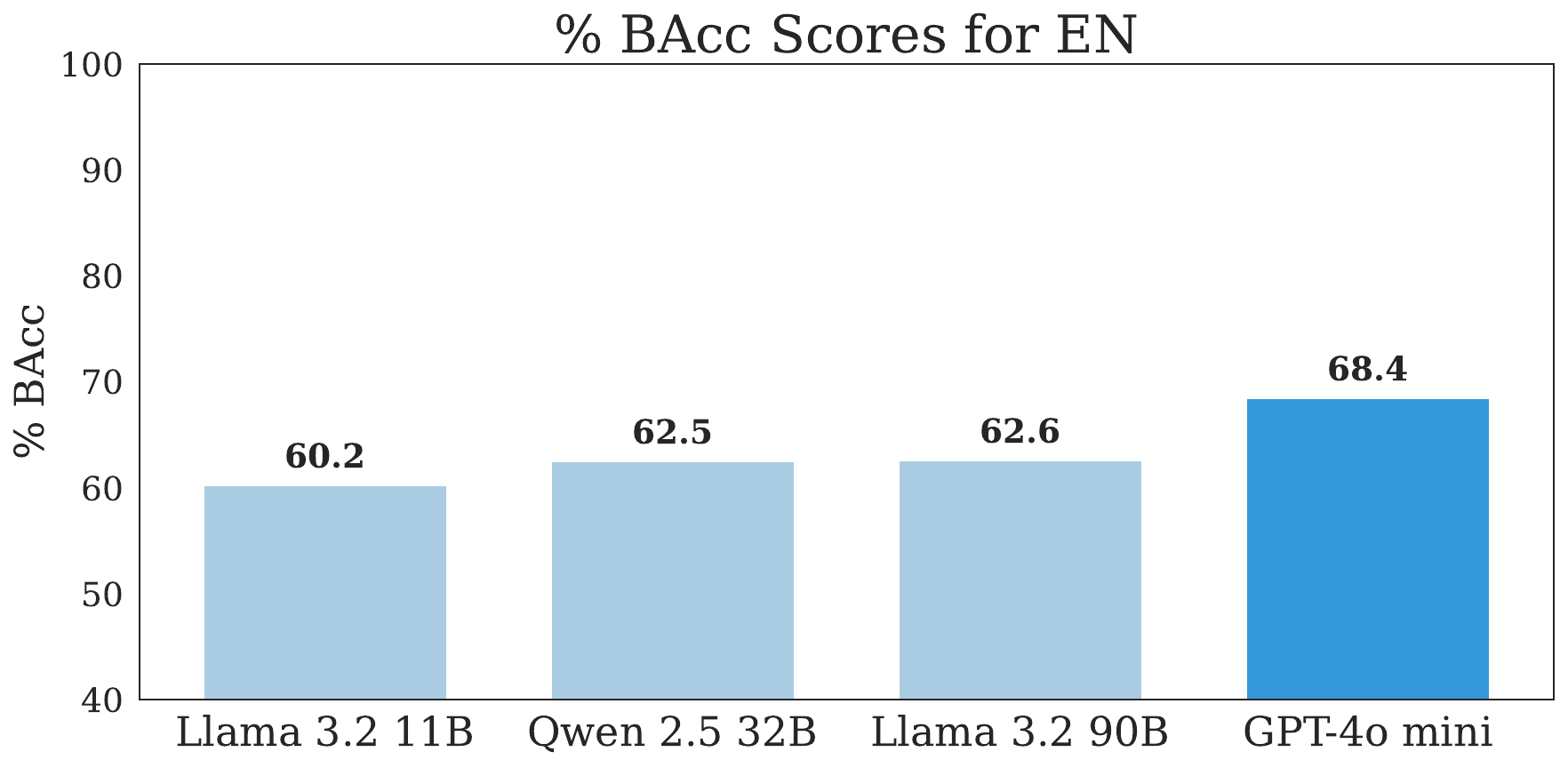} \\
    \includegraphics[width=0.48\linewidth]{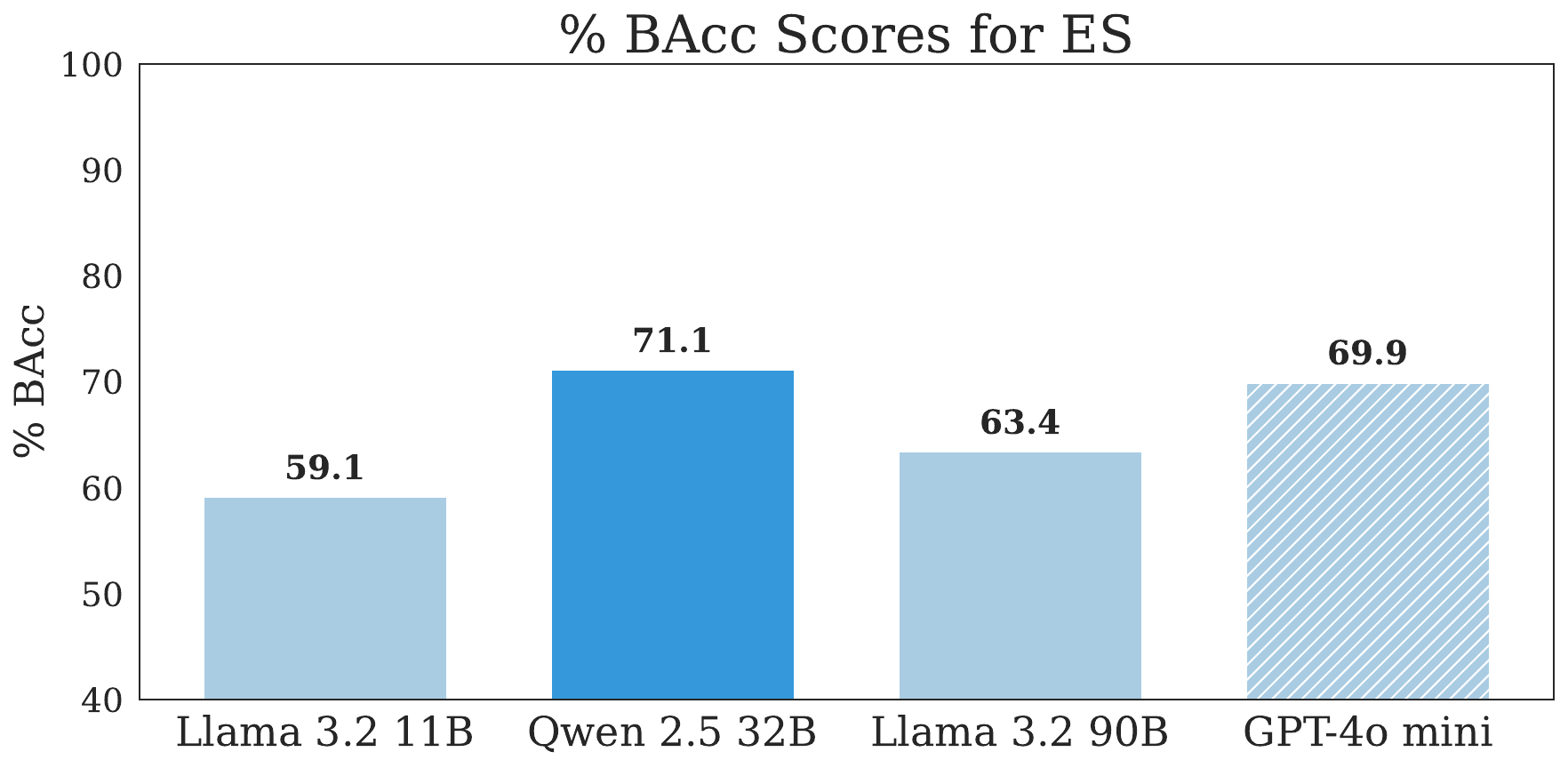} &
    \includegraphics[width=0.48\linewidth]{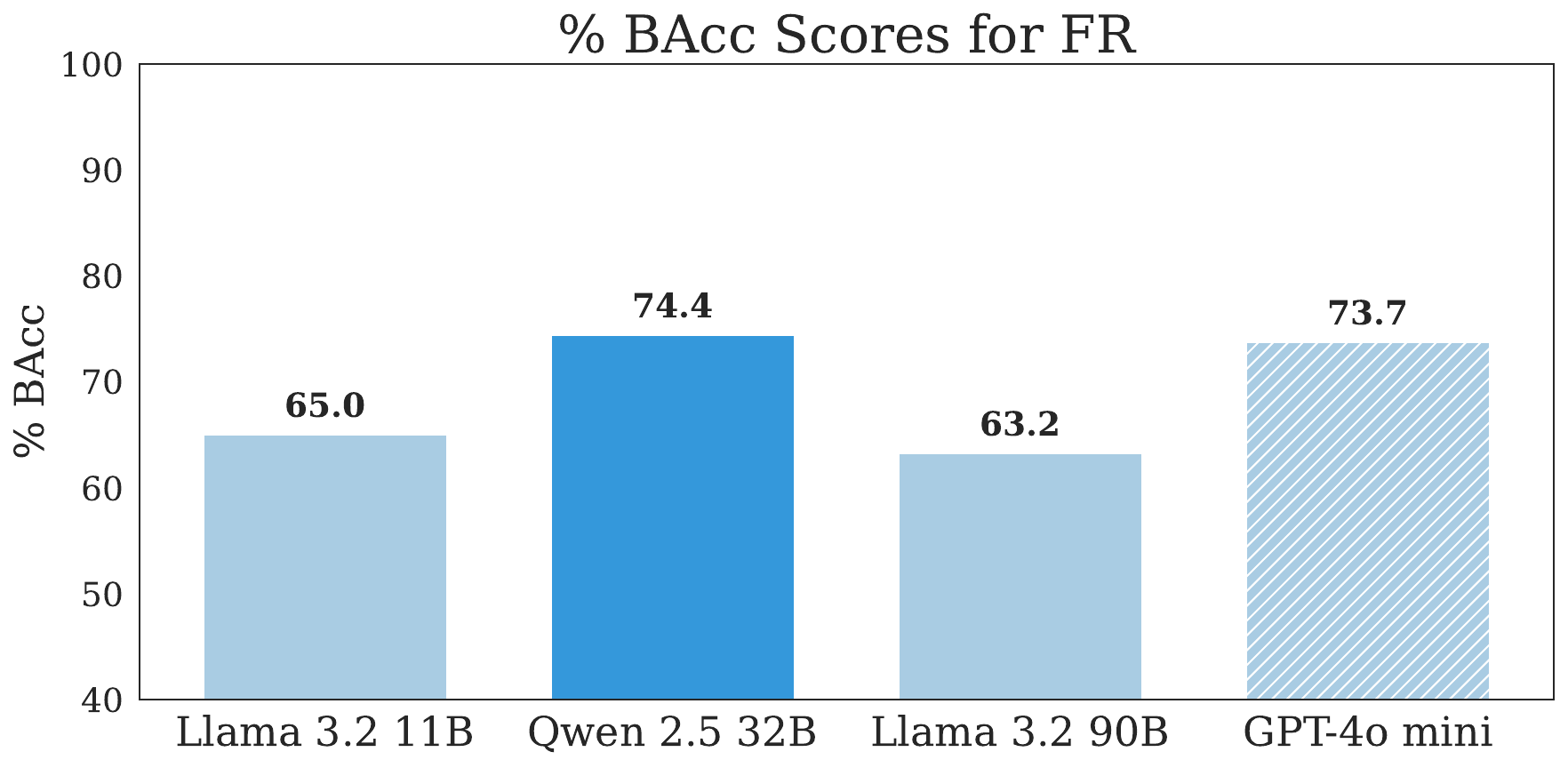} \\
  \end{tabular}
  \includegraphics[width=0.48\linewidth]{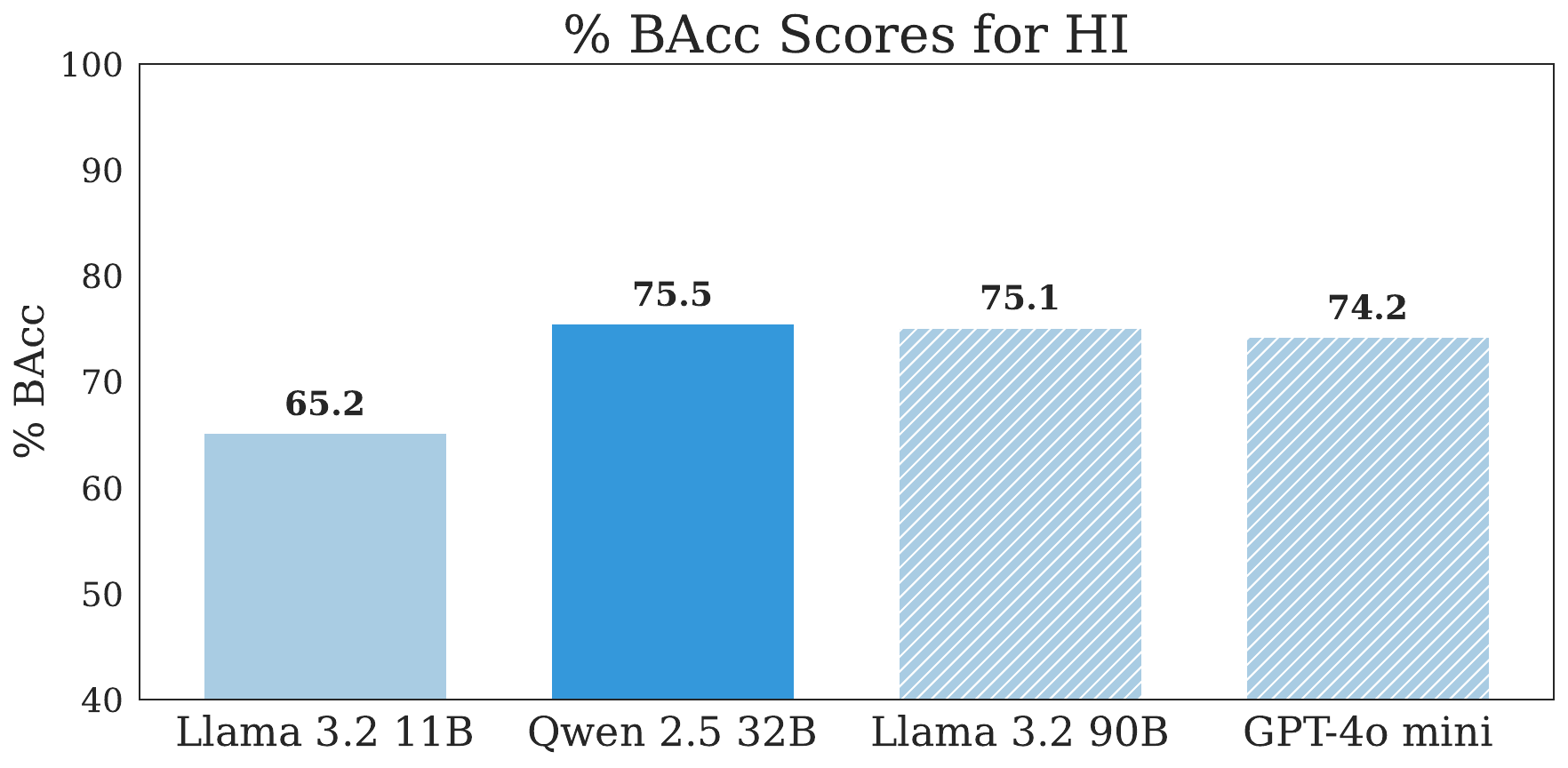}

  \caption{Reference baselines for the monolingual task  on coarse-grained faithfulness. Balanced Accuracy (BAcc) of the automatic evaluators using various LLMs across five languages: EN, DE, ES, FR and HI. Each plot compares the performance of four models: Llama 3.2 11B, Qwen 2.5 32B, Llama 3.2 90B, and \gptfomini using AG + COT prompt. The best-performing model for each language is highlighted with a darker blue bar. Bars with diagonal hatching indicate results not statistically different from the best (p > 0.05). }
  \label{fig:bacc_per_lang}
\end{figure*}

\subsection{Experimental Results}
Our benchmark enables systematic evaluation of different approaches to automated faithfulness evaluation. To illustrate this, we examine how the benchmark can surface the effectiveness of various automatic evaluation models and prompting strategies.

Table \ref{tab:avg_prompt_bacc} demonstrates the benchmark's ability to compare different prompting approaches across languages. The benchmark reveals consistent patterns, showing how different prompt designs impact evaluation quality. As expected, adding a reasoning step (COT) improves over zero-shot prompting. In addition, adding annotation guidelines (AG) helps align automated evaluators with human judgments across all languages. 
Comparing the two best models \gptfomini and \qwenthrirtytwob, \qwenthrirtytwob excels in the zero-shot and COT setups, showcasing higher ``out-of-the-box'' alignment with human judgements. \gptfomini achieves similar performance once the annotation guidelines are added to the prompt. 


Figure ~\ref{fig:bacc_per_lang}, shows the performance per language of various automatic evaluators with a fixed prompt (AG + COT), which allows us to select the best model for each language. We observe that \gptfomini performs best in English. For the rest of the languages \qwenthrirtytwob performs the best however, the results are not statistically different from \gptfomini. Our benchmark also provides users with the capability to conduct detailed, fine-grained analyses of model performance across various dimensions of faithfulness. The breakdown of automatic evaluation performance by error type is shown in Appendix~\ref{appendix:fine_grained_analysis}.

\label{sec:prompt_perf}

\looseness=-1

\section{Conclusions}
We introduced a high-quality and challenging multilingual end-to-end meta-evaluation benchmark for RAG (MEMERAG). Our carefully designed flow-chart-based annotation achieved a high inter-annotator agreement rate supporting the reliability of the benchmark. The introduced MEMERAG dataset opens the door for multiple application scenarios, including but not limited to the demonstrated cases, i.e. prompt selection and model selection.

For the meta-evaluation setup, we develop and compare various LLMs-as-a-judge and observe that automatic evaluators performance varies across the languages, influenced by language characteristics, native-question complexity and LLM generation nuances. These variations underscore the importance of our testbed, which demonstrated consistent results when comparing the prompting techniques (COT+guidelines > COT > zero-shot) and provides a foundation for developing better multilingual evaluators.


\section{Limitations}
Due to time and cost constraints, our annotations and experiments are limited in terms of prompting techniques, LLMs we experimented with and languages we annotated. Nevertheless, we diversified our LLMs across size and ``openness'' while the languages represent two families and low and high resource ones. In addition, there exists in the literature fine-tuned factuality evaluators for English, though we expect those to not work as well on non-English languages. Another method is to approximate factuality through entailment tasks (i.e. XNLI dataset) though such methods were shown (for English) to be inferior to multi-task training and distillation and data augmentation from LLMs~\cite{tang_minicheck_2024}. Fine-tuning multilingual evaluators and examining transfer learning across languages is interesting but is left for future work that can leverage our dataset for this purpose.

As we advocate for a native testing approach, the questions across the languages are not parallel, which could introduce a dimension of different questions and LLM generations complexities across the different language test data. The data we collected presents different challenges which are captured according to our fine grained error labels (Table~\ref{tab:mistake_analysis}). Future work could balance the challenges and complexities by collecting data for specific challenging phenomena. Note that this balancing is not straightforward, it can be done on the question side though this is insufficient as it does not control for the answer complexity. Controlling for the answer complexity is a challenging problem as the answer side is model generated (one method is to generate many answers and select for certain phenomena with human in the loop which is costly).

\bibliography{custom}

\newpage
\appendix

\begin{figure*}[!h]
  \centering
  \includegraphics[width=\textwidth]{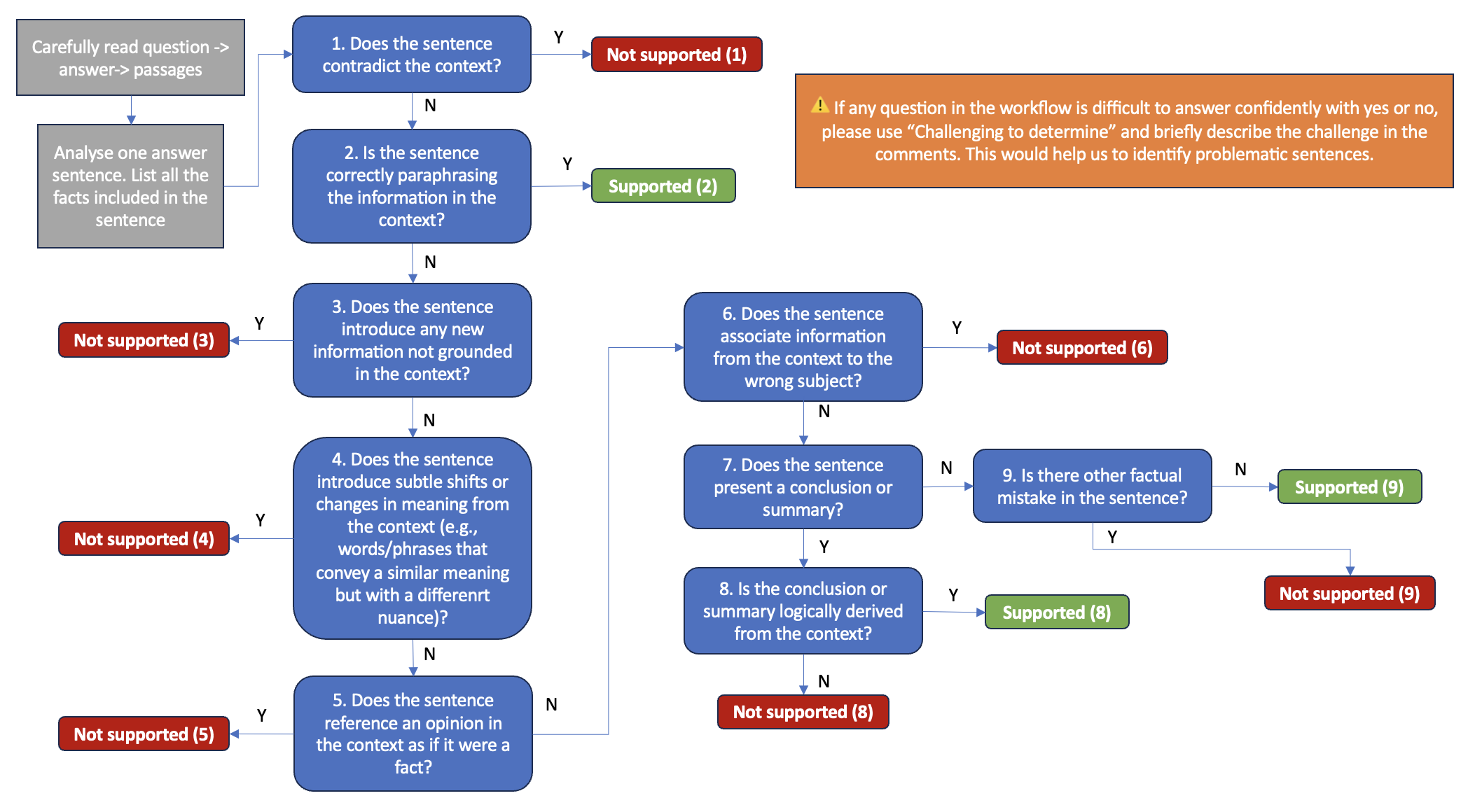}
  \caption{Annotation guideline for faithfulness labelling by human}
  \label{fig:annotation_guide}
\end{figure*}

\begin{figure*}[!tbp]
  \centering
  \includegraphics[width=\textwidth]{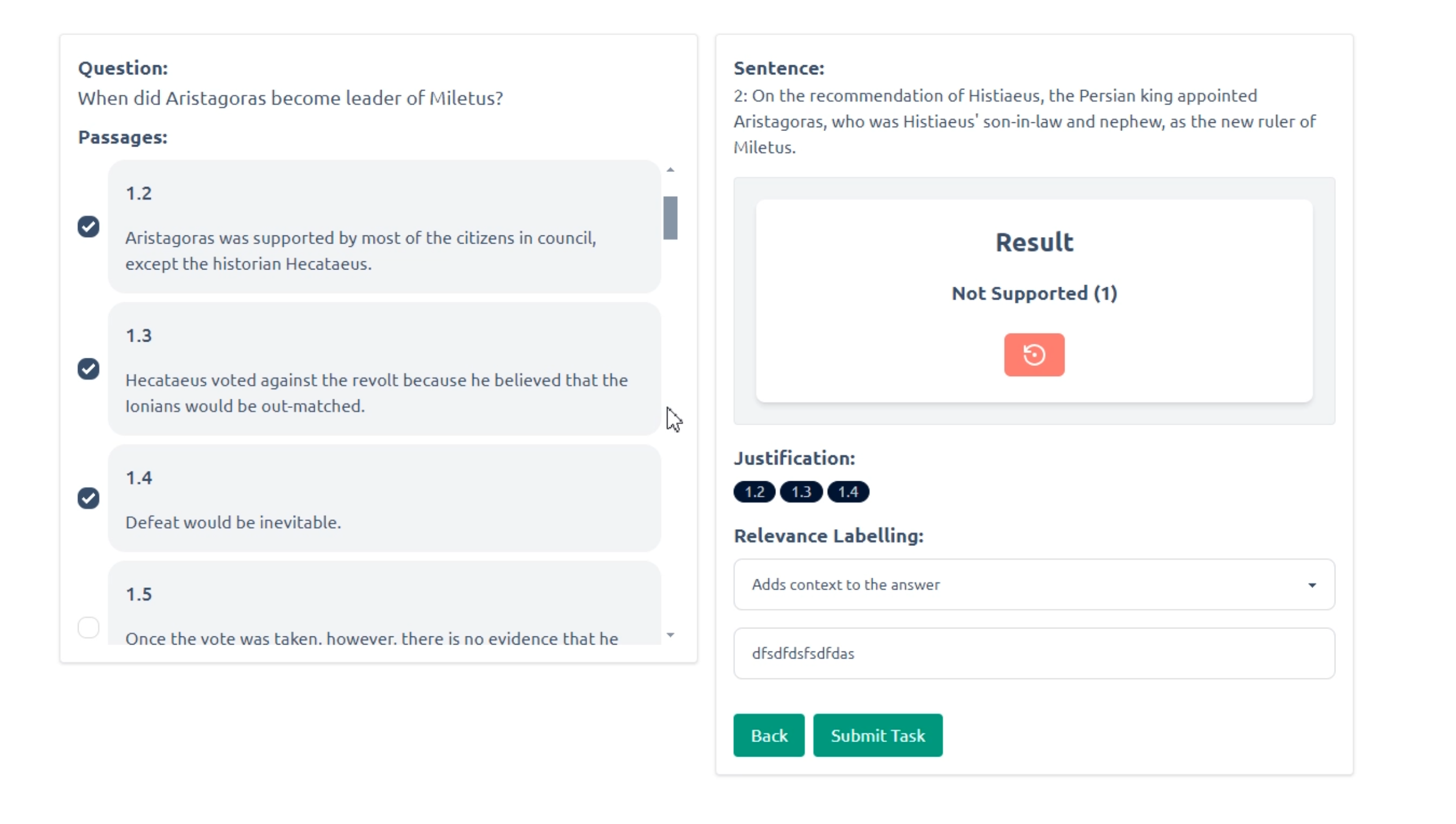}
  \caption{User interface used by human annotators for labelling faithfulness and relevance.}
  \label{fig:labelling_ui}
\end{figure*}

\section{Human Annotation Guidance}
\label{appendix:annotation_guideline}

Before conducting large-scale annotations, we conducted a pilot with 10 English RAG outputs and 3 annotators.
We asked annotators to evaluate faithfulness at the sentence level either as \textit{Supported} or with a subset of the factuality mistakes typology in~\citet{tang_tofueval_2024} (developed for summarization): \textit{Contradiction}, \textit{Hallucination}, \textit{Mis-referencing}, \textit{Nuance meaning shift}, \textit{Opinion stated as fact}, \textit{Wrong reasoning}, to which was an \textit{Other mistake} label was added to account for unforeseen mistakes in the RAG setting.
As this led to very low IAA, we conducted another round reducing the non-supported labels to \textit{Stating opinion as fact}, \textit{Drawing wrong conclusions}, \textit{Other mistake} but this still resulted in low IAA (Gwet's AC1 0.45).

Upon careful analysis of the annotator disagreements in the pilot and further rounds of calibration, we developed the annotation workflow shown in Figure~\ref{fig:annotation_guide}.
The key improvements were:
(i) add a \textit{Challenging to determine} label,
(ii) have two levels of labels, a coarse-grained level (\textit{Supported}, \textit{Not supported}, \textit{Challenging to determine}) and a fine-grained level for precision on the mistakes, 
(iii) enforce annotators to follow a specific reasoning with a flow chart,
(iv) use numbers for the fine-grained level rather than labels which could be misinterpreted.
Those guidelines allowed to reach significantly higher IAA on faithfulness (Gwet's AC1 0.81 coarse-grained).
Likewise we iterated on the relevance labels, starting from \textit{Must have}, \textit{Nice to have}, or \textit{Irrelevant} and converging to the more explicit \textit{Directly answers the question}, \textit{Adds context to the answer}, or \textit{Unrelated to the question}.
This also increased the IAA significantly. The screenshot of the user interface used by human annotators is shown in Figure ~\ref{fig:labelling_ui}.

\section{IAA and Extended Dataset}
\label{appendix:iaa_and_ext_dataset}

\begin{table}[h]
    \centering
    \normalsize{
    \begin{tabular}{ccclll}
\multicolumn{6}{c}{\textbf{\dataname}} \\
    \hline
    \multirow{3}{*}{\textbf{Lang}} & \multirow{3}{*}{\textbf{\#S}} & \multicolumn{4}{c}{\textbf{IAA (Gwet's AC1)}} \\
     &   & \multicolumn{2}{c}{\textbf{Faithfulness}} & \multicolumn{2}{c}{\textbf{Relevance}} \\
    &   & \multicolumn{1}{c}{\textbf{Coarse}} & \multicolumn{1}{l}{\textbf{Fine}} & \textbf{Coarse} & \textbf{Fine} \\ 
    \hline
    EN  & 13 & 0.93 & 0.63 &    1 & 0.92 \\
    DE  & 31 & 0.84 & 0.47 & 0.95 & 0.91 \\
    ES  & 20 & 0.91 & 0.3  &    1 & 0.93 \\
    FR  & 23 & 0.89 & 0.58 & 0.93 & 0.92 \\
    HI  & 17 & 0.89 & 0.18 &    1 & 1    \\\hline
  \end{tabular}
  }
    \caption{Inter-annotator agreement (IAA) on the faithfulness and relevance dimensions with 3 annotators (\dataname subset), for coarse-grained and fine-grained levels. Annotations are at the answer sentence level (\#S number of answer sentences is provided).
    }
    \label{tab:iaa_appendix}
\end{table}


\begin{table}[h]
    \centering
    \normalsize{
\begin{tabular}{ccclll}
\multicolumn{6}{c}{\textbf{\datanameext}} \\
\hline
\multirow{3}{*}{\textbf{Lang}} & \multirow{3}{*}{\textbf{\#S}} & \multicolumn{4}{c}{\textbf{IAA (Gwet's AC1)}}                                      \\
                               &                               & \multicolumn{2}{c}{\textbf{Faithfulness}} & \multicolumn{2}{c}{\textbf{Relevance}} \\
                               &                               & \textbf{Coarse}      & \textbf{Fine}      & \textbf{Coarse}     & \textbf{Fine}    \\ \hline
EN & 226 & 0.83 & 0.41 & 1.00 & 0.9  \\
DE & 272 & 0.75 & 0.47 & 0.92 & 0.79 \\
ES & 276 & 0.91 & 0.39 & 1.00 & 0.97 \\
FR & 370 & 0.72 & 0.53 & 0.99 & 0.8  \\
HI & 208 & 0.91 & 0.33 & 0.98 & 0.92 \\ \hline
\end{tabular}
  }
    \caption{Inter-annotator agreement (IAA) on the faithfulness and relevance dimensions with 5 annotators (\datanameext), for coarse-grained and fine-grained levels. Annotations are at the answer sentence level (\#S number of answer sentences is provided).
    }
    \label{tab:iaa_appendix_ext}
\end{table}

As explained in Section~\ref{sec:annotations_process}, the IAA on the \dataname dataset was measured by triple annotation on a random subset of 10 questions per language.
Detailed IAA metrics, at both coarse- and fine-grained levels, can be found in Table~\ref{tab:iaa_appendix}.
We observe that IAA is high for faithfulness at coarse-grained level, and for relevance at both levels.
This prompted us to 1) focus the experiments described in this paper to those dimensions with high IAA, and 2) investigate the low IAA for fine-grained faithfulness.

The disagreements highlighted by low IAA are due to ambiguity in the annotation guidelines.
The fact that such ambiguity remained after significant work on refining the guidelines and guiding annotators with UI elements hints at the irreducible subjectivity of the task.
Subjectivity manifests itself as a non-trivial probability distribution of the annotation label obtained by drawing a random human annotator for a given question.
It would be of interest in the development of automated annotators to compare this distribution to that of the annotations of a stochastic LLM-based annotator.

To investigate those aspects, we created an extended dataset, called \datanameext.
The dataset is composed of 150 question-context pairs per language, extracted similarly to \dataname with the process described in Section~\ref{sec:dataset_construction}.
The questions are disjoint between the two datasets, except for German where there is some overlap due to the limited number of questions in MIRACL.
The annotation process then differed in two ways: 1) annotations were obtained for all question from 5 expert annotators, and 2) The \textit{Challenging to determine} label was not an option for annotators to force meaningful labels even at the cost of disagreements.
IAA metrics for the \datanameext dataset can be found in Table~\ref{tab:iaa_appendix_ext}.
Those results are in line with the IAA observed in \dataname, which confirms its quality.
Detailed investigation of the disagreements between human annotators, and comparison to automated annotators are left as future work.
We release this extendended dataset along with the main dataset to promote related work by the research community.

\begin{table}
    \centering
    \normalsize{
    \begin{tabular}{lcccc}
    \hline
    \multirow{2}{*}{\textbf{Lang}} & \multirow{2}{*}{\textbf{\#Q}} & \multicolumn{2}{c}{\textbf{Answer}} & \textbf{Context}  \\
                                   &                               & \textbf{\#S}   & \textbf{Avg. \#W}  & \textbf{Avg. \#W} \\ \hline
    EN    & 150 & 226  & 27.5 & 591.4 \\
    DE    & 150 & 272  & 30.0 & 456.8 \\
    ES    & 150 & 276  & 39.1 & 490.8 \\
    FR    & 150 & 370  & 48.1 & 453.2 \\
    HI    & 150 & 208  & 23.0 & 576.6 \\ \hline
    Total & 750 & 1352 &      &       \\ \hline
    \end{tabular}
  }
    \caption{General statistics of the \datanameext dataset. \#Q: number of questions, \#S: number of sentences, \#W: number of words.
    Each answer is annotated at the sentence level, leading to 1,352 total sentences annotated by experts for faithfulness and relevance.
    }
    \label{tab:dataset_stats_ext}
\end{table}



\section{Detailed numbers on the \dataname dataset}
\label{appendix:memereg-detail}
\begin{table*}[htbp]
    \centering
    \small{
    \begin{tabular}{lccccc}
    \hline
    \textbf{Model} & \textbf{EN} & \textbf{DE} & \textbf{ES} & \textbf{FR} & \textbf{HI}\\
    \hline
    LLM-A (615) & & & & & \\
    \textit{Faithfulness} & 52.7 / 43.8 / 3.5 & 74.3 / 17.1 / 8.6 & 58.6 / 41.4 / 0 & 59.6 / 40.4 / 0 & 73.2 / 26.8 / 0\\ 
    \textit{Relevance} & 42.5 / 51.7 / 5.8 & 60.9 / 37.4 / 1.7 & 42.3 / 54.5 / 3.2 & 48 / 50.3 / 1.7 & 46.8 / 41.3 / 11.9 \\
    LLM-B (315) & & & & & \\
    \textit{Faithfulness} & {76.1} / 17.9 / 6 & {88.4} / 11.6 / 0 & {88.2} / 11.8 / 0 & 58.6 / 41.4 / 0 & {84.9} / 15.1 / 0 \\
     \textit{Relevance} & 67.1 / 31.5 / 1.4 & 72.7 / 22.1 / 5.2 & {75} / 22.2 / 2.8 & {87.1} / 12.9 / 0 & 80.7 / 10.5 / 8.8 \\
    LLM-C (315) & & & & & \\
    \textit{Faithfulness} & 61.4 / 35.1 / 3.5 & 69.8 / 30.2 / 0 & 68.1 / 31.9 / 0 & {80.3} / 19.7 / 0 & 80 / 20 / 0 \\
     \textit{Relevance} & {86.9} / 11.5 / 1.6 & {78.3} / 20.3 / 1.4 & 69.3 / 28 / 2.7 & {87.1} / 10 / 2.9 & 85.9 / 9.4 / 4.7\\
    LLM-D (743) & & & & & \\
    \textit{Faithfulness} & 67.3 / 30.8 / 1.9 & 61.4 / 38 / 0.6 & 61 / 35.7 / 3.3  & 59.8 / 40.2 / 0 & 63 / 34.6 / 2.5\\
     \textit{Relevance} & 70.3 / 29.7 / 0 & 41.5 / 33 / 25.5 & 33 / 52.3 / 14.7 & 54.8 / 26.1 / 19.1 & 57.1 / 24.2 / 18.7\\
    LLM-E (334) & & & & & \\
    \textit{Faithfulness} & 77.2 / 21.1 / 1.7 & 73.9 / 26.1 / 0 & 69.7 / 29 / 1.3 & 59.2 / 39.5 / 1.3 & 73.3 / 26.7 / 0\\
     \textit{Relevance} & 80.3 / 19.7 / 0 & 75.4 / 17.4 / 7.3 & 63.4 / 36.6 / 0 & 73.8 / 25 / 1.2 & {90.6} / 9.4 / 0 \\\hline
  \end{tabular}
    \caption{Percentage of sentences labelled as \textit{Supported/Not Supported/Challenging to determine} for the faithfulness, and as \textit{Directly answers the question/Adds context to the answer/Unrelated to the question} for the relevance, per language and model. The total number of sentences generated by the model across all languages is given in parenthesis besides the model name.
    }
    }
\end{table*}

In this section, we look at the detailed statics of faithfulness and relevance in the MEMERAG dataset itself. Overall, the faithfulness and relevance in the dataset variate a lot from one language to another language and from one generator model to another.
There is usually a range of 10-20 percentage points between the language with lowest percentage and the language with highest percentage of supported sentences. This supports our assumption of variable behaviour/performance across different languages. 
 
Given the large variation, the percentage of supported answers is larger than that of unsuported answers. Similarly, the ratio of relevant answers is in general larger than that of other relevance buckets (except for ES with LLM-D; EN, ES and FR with LLM-A).
Interestingly, when analysing the vairous results per model and language, we did no find general pattern. This is counter intuitive to our assumption of having more supported sentences for English language. Surprisingly, even though Hindi is a low-resource language the models are producing 70\% supported sentences, except for LLM-D that produced only around 62\% supported sentences.

\section{Fine grained automatic annotation analysis}
\begin{figure*}[!htbp]
  \centering
  \includegraphics[width=\textwidth]{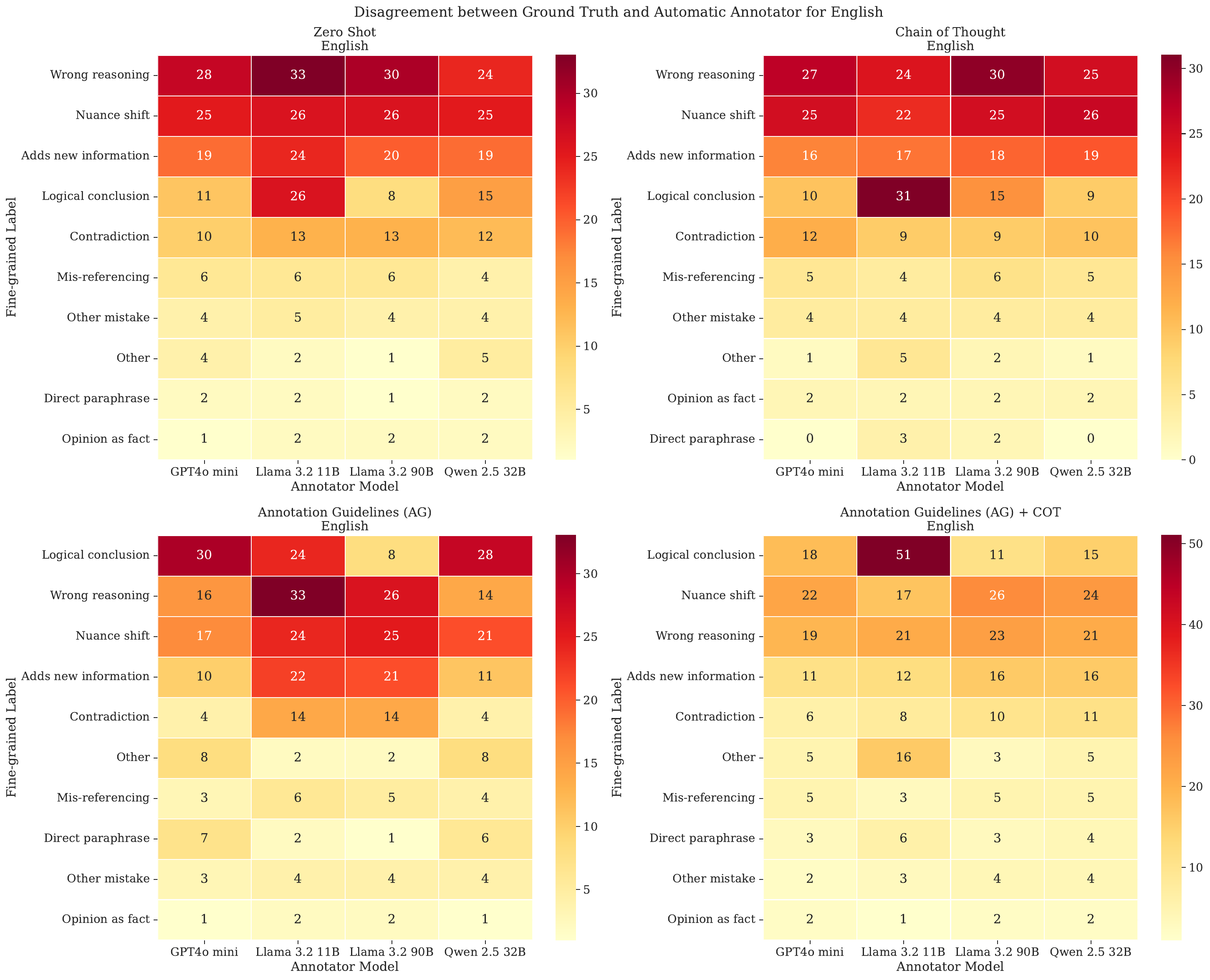}
  \caption{Fine-grained faithfulness errors when automatic evaluator disagrees with ground truth.}
  \label{fig:zero_shot_en_heat_map}
\end{figure*}

Figure \ref{fig:zero_shot_en_heat_map} shows a heatmap representing the failure modes of automatic evaluators. We observe that improving the prompting strategy from ZS to AG + COT, reduces automatic evaluation errors across all error categories, except for Llama 3.2 11B, where the model makes more errors in the ``Logical conclusion category''. We also observe that ``Wrong reasoning'', ``Nuance shift'' and ``Logical conclusion'' are the top error categories for all the models tested. Future work could explore prompting or fine-tuning techniques designed to handle specific error types.

\section{Statistical Significance Test Details}
\label{sec:stat_test}
In Table~\ref{tab:avg_prompt_bacc} and Figure~\ref{fig:bacc_per_lang}, we aim to determine whether the scores (in terms of BAcc) for the best prompt and best model was significantly different from the other scores in the table. To test statistical significance, we use permutation test \cite{good2013permutation}, a non-parametric method for comparing two related samples. The null hypothesis for this test suggests that there is no significant difference between the performance of the best-performing prompt/models and the other prompts/models, while the alternative hypothesis suggests a significant difference exists. We consider $\alpha = 0.05$ as significance level. Consequently, when p > 0.05, we are not able to reject the null hypothesis, indicating that prompts and models have similar performance.






\section{Prompts used for dataset construction}
\label{sec:prompts_dataset_construction}
In section we describe the various prompts used in our pipeline. Out prompts are written as  Jinja2\footnote{\url{https://jinja.palletsprojects.com/en/stable/}} templates.
\subsection{Time-dependent answer filtering}
\label{prompt:time_dependent}
During internal pilots, we identified answers that relied on current date / current affairs that could be challenging to determine their faithfulness. For example, to the question "How old is Yann LeCun?" is the answer correct if the LLM uses the time when it was trained? To avoid such cases we used Llama3 70B to filter out those cases using the prompt shown below:

\begin{tcolorbox}[colback=gray!10, colframe=gray!50, title= Time-dependent Answer filtering]
\small{
\label{prompt:time_dependent}
{\ttfamily You are an NLP assistant that helps to identify if a question requires to know  when is today (day, or month, or year), current affairs, or up-to-date  information. Give the step by step of how to answer the question in between the labels <rationale></rationale>. Then verify if the steps included to know 
any information about the current time and give your answer in between the 
tags: <label></label>. Please only use 'yes' or 'no' in your final answer.  You will be given the question in \{\{language\}\}. 

Provide your rationale and label in English. 
}
}
\end{tcolorbox}

\subsection{Highlighting relevant segments}
We highlight relevant statements in the passage to help annotators focus on important parts of the context. We prompt Llama 3 70B, to predict all the statements that support the passage. We use temperature equal to 0.1 in this step.

\subsection{Answer generation prompt}
\label{sec:prompts_answers}
We generate the answers using the same prompt across all models and languages. All the instructions were given in English, while the context and question were given in the testing language. For all models we set the temperature to 0.1 and the maximum number of token to 1000.\\

\begin{tcolorbox}[colback=gray!10, colframe=gray!50, title= Highlighting Relevant Segments]
\small
\label{sec:passage_highlight}
{\ttfamily You are an agent that verifies if sentences are supported by a given context. \\
You will be given a context made of several passages referred as "Passage" split 
into sentences, each with a numeral identifier, and each referred as "Sentence". 
Your task is to determine if the sentence is supported by some of the passages  
(using label 1)  or is not supported (using label 0). Please follow this process: \\ \\
(1) Explain why the sentence is supported or not supported, please write your reasoning in between the tags <rationale></rationale>. If a sentence is supported, 
list the number of the sentences in the passage or several passages that supported 
it. If a sentence is not supported, explain why is not supported.\\  \\
(2) Write the  final label for the sentence in between the tags <label></label>. \\ \\
(3) Write the id  of the supporting sentences from passages in between the tags  <references></references> separated by comma. Remember to use only 0 or 1 for label. \\

\begin{verbatim}
### Context
{% for passage in passages -%}
Passage {{loop.index}}: 
{% set outer_loop = loop %}
{% for sent in passage.text %}
    {{outer_loop.index}}.{{loop.index}}: 
    {{sent|safe}}
{% endfor %}

{% endfor %}

### Sentence: 
{{sentence|safe}}
\end{verbatim}

}

\end{tcolorbox}

\begin{tcolorbox}[colback=gray!10, colframe=gray!50, title= Answer Generation Prompt]
\small
\label{prompt:answer_generation}
{\ttfamily You are an NLP assistant whose purpose is to answer a question based on given passages. The passages may or may not help answer the question. You will need to 
provide the answer based only on the passages. The answer must be in {{language}} 
without fail. Be concise and direct without referring to passages in the answer. 
Avoid expressions such as "According to the passages" or "Based on the passages". \\

\begin{verbatim}
{% for passage in passages -%}
    - {{passage.text|safe}}
{% endfor %}

Question: {{question}} 

Please answer directly the question 
above in {{language}}.
\end{verbatim}
}

\end{tcolorbox}

\section{Prompts for automated evaluation}
\label{sec:eval_prompts}
The four prompts evaluate answer faithfulness to source passages with increasing complexity: Zero-shot (ZS) provides basic supported/not-supported classification, Chain of Thought (COT) adds explicit reasoning steps, Annotation Guidelines (AG) includes detailed evaluation criteria, and AG+COT combines detailed guidelines with reasoning steps. All prompts output their final classification in <answer> tags.

\begin{tcolorbox}[colback=gray!10, colframe=gray!50, title= Zero-shot Prompt (ZS)]
\label{sec:prompt_zs}
\small
{\ttfamily You are an automatic annotator tasked with determining whether a given answer is grounded in the list of provided evidence passages. Your role is to carefully analyze the relationship between the answer and the evidence, and then classify the answer as either "Supported" or "Not 
Supported". Provide your answer directly in <answer> </answer> tag. \\

Evidence Passages:
\begin{verbatim}
{% for passage in context %}

  .  {{loop.index}}: {{passage.text}}

{% endfor %}

Answer: {{answer segment}}

\end{verbatim}

Now provided your label directly as ''Supported'' or ''Not Supported''.
}

\end{tcolorbox}


\subsubsection*{}
\begin{tcolorbox}[colback=gray!10, colframe=gray!50, title= Chain of thought (COT)]
\label{sec:prompt_cot}
\small
{\ttfamily You are an automatic annotator tasked with determining whether a given answer is grounded in the list of provided evidence passages. Your role is to carefully analyze the relationship between the answer and the evidence, write your reasoning in between the tags <rationale></rationale> then classify the answer as either ``Supported'' or ``Not Supported'' and write answer in <answer> </answer> tag \\

Evidence Passages:
\begin{verbatim}

{% for passage in context %}

    {{loop.index}}: {{passage.text}}

{% endfor %}

Answer: {{answer segment}}

\end{verbatim}
Now provided your label directly as "Supported" or "Not Supported".

}

\end{tcolorbox}
{\vfill\color{white}.}

\subsubsection*{}

\begin{tcolorbox}[colback=gray!10, colframe=gray!50, title= Annotation Guidelines (AG)]
\small
\label{sec:prompt_ag}
{\ttfamily Given a set of evidence passages, and an answer, determine if the answer is fully supported by the evidence passages or not. Analyze each sentence of the answer carefully and verify that all information it contains is explicitly stated in or can be directly inferred from the evidence passages. \\\\

Output "Not Supported" if ANY of the following are true: \\

The answer contains any information not explicitly stated in or directly inferable from the passages. \\\\
The answer contradicts any information in the passages. \\\\
The answer introduces any new information not found in the passages.\\\\
The answer misrepresents or inaccurately paraphrases information from the passages. \\\\
The answer draws conclusions not logically supported by the given information.\\\\
The answer changes the level of certainty, specificity, or nuance from what is expressed in the passages.\\\\
The answer does not directly address the specific aspect asked about in the question.\\\\
The answer conflates or misrepresents separate pieces of information when summarizing multiple passages. \\\\

Output Supported otherwise.

Provide this determination without any additional explanation in <answer></answer> tags. Analyze thoroughly but output only the single-word label. \\

Evidence Passages:
\begin{verbatim}
{% for passage in context %}

    {{loop.index}}: {{passage.text}}

{% endfor %}

Answer: {{answer segment}}
\end{verbatim}\\

Now provided your label directly as "Supported" or "Not Supported".

}

\end{tcolorbox}
\subsubsection*{}
\begin{strip}
\begin{tcolorbox}[colback=gray!10, colframe=gray!50, title= Annotation guidelines with Chain of thought (AG + COT)]
\label{sec:prompt_ag_cot}
{\ttfamily Given a set of evidence passages, and an answer, determine if the answer is fully supported by the evidence passages or not. Analyze each sentence of the answer carefully and verify that all information it contains is explicitly stated in or can be directly inferred from the evidence passages. \\

Output \textbf{"Not Supported"} if ANY of the following are true:\\

The answer contains any information not explicitly stated in or directly inferable from the passages.\\

The answer contradicts any information in the passages. \\

The answer introduces any new information not found in the passages. \\

The answer misrepresents or inaccurately paraphrases information from the passages. \\

The answer draws conclusions not logically supported by the given information. \\

The answer changes the level of certainty, specificity, or nuance from what is expressed in the passages. \\

The answer does not directly address the specific aspect asked about in the question. \\

The answer conflates or misrepresents separate pieces of information when summarizing multiple passages. \\

Output \textbf{Supported} otherwise.

write your reasoning in between the tags \textbf{<rationale></rationale>} and Provide your final answer in \textbf{<answer></answer>} tags. \\

Evidence Passages:
\begin{verbatim}
{% for passage in context %}
    {{loop.index}}: {{passage.text}}
{% endfor %}
Answer: {{answer segment}}
\end{verbatim}
Now provided your label directly as "Supported" or "Not Supported".
}
\end{tcolorbox}
\end{strip}

\section{More Results on Meta-Evaluators}
\begin{table*}[!t]
\begin{tabular}{@{}cccccc@{}}
\toprule
\multicolumn{1}{l}{} & EN                        & DE               & ES               & FR               & HI               \\ \midrule
GPT4o mini           & \textbf{68.37 $\pm$ 2.84} & 73.58 $\pm$ 2.86 & 69.84 $\pm$ 2.26 & 73.74 $\pm$ 2.29 & 74.10 $\pm$ 2.61 \\
Qwen 2.5 32B & 62.45 $\pm$ 3.00 & \textbf{76.79 $\pm$ 2.76} & \textbf{71.09 $\pm$ 1.98} & \textbf{74.38 $\pm$ 2.28} & \textbf{75.41 $\pm$ 2.72} \\
Llama 3.2 90B        & 62.60 $\pm$ 2.73          & 63.11 $\pm$ 2.70 & 63.36 $\pm$ 2.09 & 63.17 $\pm$ 1.97 & 75.05 $\pm$ 2.87 \\
Llama 3.2 11B        & 60.29 $\pm$ 2.84          & 59.91 $\pm$ 3.13 & 59.02 $\pm$ 2.51 & 65.02 $\pm$ 2.46 & 65.22 $\pm$ 2.75 \\ \bottomrule
\end{tabular}
\caption{Balanced Accuracy for five languages  and four LLMs using AG+COT prompting strategy. Standard errors are calculated using 1000 Bootstrap iterations.}
\label{tab:avg_llm_bacc_with_se}
\end{table*}

\begin{table*}[!htbp]
\begin{tabular}{@{}l
c
cccc@{}}
\toprule
\textbf{Prompt} & \textbf{EN}      & \textbf{DE}               & \textbf{ES}      & \textbf{FR}      & \textbf{HI}      \\ \midrule
ZS              & 57.22 {$\pm$} 2.14 & 61.54 $\pm$ 1.98          & 56.94 $\pm$ 1.26 & 60.00 $\pm$ 1.39 & 68.96 $\pm$ 2.22 \\
COT             & 60.71 $\pm$ 2.34 & 63.83 $\pm$ 2.15          & 60.61 $\pm$ 1.44 & 63.85 $\pm$ 1.53 & 69.68 $\pm$ 2.42 \\
AG              & 62.72 $\pm$ 2.24 & \textbf{68.87 $\pm$ 2.27} & 63.61 $\pm$ 1.36 & 68.32 $\pm$ 1.53 & 70.20 $\pm$ 2.02 \\
AG + COT & \textbf{63.42 $\pm$ 2.44} & 68.35 $\pm$ 2.37 & \textbf{65.83 $\pm$ 1.64} & \textbf{69.08 $\pm$ 1.71} & \textbf{72.44 $\pm$ 2.07} \\ \bottomrule
\end{tabular}
\caption{Balanced Accuracy for five languages using four different prompting strategies, averaged across four LLMs (\gptfomini, Llama 3.2 90B, Llama 3.2 11B, and \qwenthrirtytwob).  Standard errors are calculated using 1000 Bootstrap iterations.}
\label{tab:avg_prompt_bacc_with_se}
\end{table*}

\begin{table*}[!htbp]
\centering
\begin{tabular}{@{}ccS[table-format=2.1]S[table-format=2.1]S[table-format=2.1]S[table-format=2.1]@{}}
\toprule
\textbf{Language} & \textbf{Prompt} & \textbf{GPT4o mini} & \textbf{Llama 3.2 90B} & \textbf{Llama 3.2 11B} & \textbf{Qwen 2.5 32B} \\ \midrule
\multirow{4}{*}{EN} & ZS       & 59.84 & 58.00 & 51.00 & 60.07 \\
                    & COT      & 61.78 & 59.06 & 59.99 & 61.97 \\
                    & AG       & 69.95 & 59.40 & 52.97 & 68.54 \\
                    & AG + COT & 68.43 & 62.62 & 60.22 & 62.47 \\ \midrule
\multirow{4}{*}{DE} & ZS       & 61.10 & 58.60 & 54.84 & 71.89 \\
                    & COT      & 63.30 & 60.00 & 59.59 & 72.70 \\
                    & AG       & 71.99 & 68.60 & 60.64 & 74.64 \\
                    & AG + COT & 73.69 & 63.25 & 59.98 & 76.84 \\ \midrule
\multirow{4}{*}{ES} & ZS       & 53.78 & 56.76 & 53.65 & 63.65 \\
                    & COT      & 56.22 & 58.24 & 61.89 & 66.22 \\
                    & AG       & 69.73 & 59.59 & 54.32 & 70.81 \\
                    & AG + COT & 69.86 & 63.38 & 59.05 & 71.08 \\ \midrule
\multirow{4}{*}{FR} & ZS       & 59.54 & 56.31 & 55.99 & 68.45 \\
                    & COT      & 60.36 & 58.52 & 65.13 & 71.62 \\
                    & AG       & 74.47 & 62.41 & 60.28 & 76.24 \\
                    & AG + COT & 73.73 & 63.18 & 65.00 & 74.44 \\ \midrule
\multirow{4}{*}{HI} & ZS       & 72.23 & 64.33 & 66.12 & 73.27 \\
                    & COT      & 72.06 & 67.61 & 65.35 & 73.78 \\
                    & AG       & 73.78 & 67.10 & 65.52 & 74.53 \\
                    & AG + COT & 74.19 & 75.06 & 65.23 & 75.52 \\ \bottomrule
\end{tabular}
\caption{Meta evaluation results for all combination of LLMs and prompts we have tested.}
\label{tab:metaeval_all}
\end{table*}

\clearpage
\mbox{~}

\section{Fine grained Analysis}
\label{appendix:fine_grained_analysis}
\begin{table}[!t]
\centering
\small
\begin{tabular}{lr*{2}{S[table-format=2.1]}}
\toprule
\textbf{Fine-Grained Label} & \textbf{\#S} & \textbf{\gptfomini} & \textbf{\begin{tabular}[c]{@{}l@{}}Llama 3.2\\  90B\end{tabular}} \\ \midrule
Logical conclusion   & 151 & 80.13   & 87.42 \\
Direct paraphrase    & 182 & 93.41   & 97.25 \\
Other                & 2   & 50.00      & 50.00    \\ \midrule
We. Acc Sup.          &     & 87.16   & 92.54 \\ \midrule
Adds new information & 81  & 67.90    & 33.33 \\
Nuance shift         & 30  & 26.67   & 16.67 \\
Mis-referencing      & 20  & 35.00   & 20.00 \\
Contradiction        & 32  & 65.63   & 50.00    \\
Opinion as fact      & 12  & 66.67   & 25.00    \\
Other mistake        & 19  & 84.21   & 52.63 \\
Wrong reasoning      & 10  & 80.00      & 40.00    \\ \midrule
We. Acc NS.           &     & 60.29   & 33.82 \\ \midrule
BAcc                 &     & 73.73 & 63.18
\end{tabular}
\caption{Accuracy of \gptfomini and Llama 3.2 90B on various fine-grained labels, when evaluating French language using AG + COT prompt}
\label{tab:gpt_lama_fr_fine_grained}
\end{table}

To demonstrate the benchmark's capability for fine-grained analysis, Table \ref{tab:gpt_lama_fr_fine_grained} breaks down automatic evaluation performance by error type for two models: \gptfomini and Llama 3.2 90B, both evaluating French answers using AG + COT prompt. The upper section of the table shows the distribution of errors when the ground truth class is \textit{Supported}, categorized into logical conclusion, direct paraphrase, or other correct categories. The lower section details the types of mistakes made by each model for the \textit{Not supported} category. We observe that both models show a similar pattern for the \textit{Supported} category, with logical conclusions being the most common (69.8\% for \gptfomini and 76.0\% for Llama 3.2 90B), followed by direct paraphrases. For the \textit{Not supported} category, both models struggle most with detecting ``adding new information'' (32.1\% and 40.0\% of mistakes, respectively) and ``nuance shifts'' (27.2\% and 18.5\%). The ranking of error types is consistent across both models, despite their different overall BAcc (73.7\% for \gptfomini vs 63.2\% for Llama 3.2 90B see Table ~\ref{tab:metaeval_all} in Appendix), suggesting similar challenges in evaluation. The benchmark's fine-grained labeling system provides a detailed view of evaluation challenges across languages. By categorizing different types of faithfulness violations, it reveals which specific errors are harder to detect for each model, offering insights into both the strengths and limitations of automated evaluation methods.

\end{document}